\documentclass[12pt]{article}

\usepackage[utf8]{inputenc}
\usepackage{booktabs}

\usepackage{amsmath,array,graphicx}
\usepackage{amssymb} 
\usepackage{textcomp}
\usepackage{gensymb}
\usepackage{float}
\usepackage{graphicx}
\usepackage{textgreek}
\usepackage{multicol, multirow}
\usepackage[many]{tcolorbox}
\usepackage{makecell}
\usepackage{hyperref}

\usepackage[labelfont=bf]{caption}
\usepackage[textfont=it]{caption}
\usepackage[margin=0.75in]{geometry}
\usepackage[superscript]{cite}
\usepackage{tabularx}
\usepackage{subcaption}
\title{A collaborative digital twin built on FAIR data and compute infrastructure}

\author{Thomas M. Deucher, Juan C. Verduzco, \\
Michael Titus, and Alejandro Strachan}

\date{
School of Materials Engineering \\
Purdue University, West Lafayette, Indiana, 47907 USA
}

\begin{document}

\maketitle

\begin{abstract}

The integration of machine learning with automated experimentation in self-driving laboratories (SDL) offers a powerful approach to accelerate discovery and optimization tasks in science and engineering applications. When supported by findable, accessible, interoperable, and reusable (FAIR) data infrastructure, SDLs with overlapping interests can collaborate more effectively. This work presents a distributed SDL implementation built on nanoHUB services for online simulation and FAIR data management. In this framework, geographically dispersed collaborators conducting independent optimization tasks contribute raw experimental data to a shared central database. These researchers can then benefit from analysis tools and machine learning models that automatically update as additional data become available. New data points are submitted through a simple web interface and automatically processed using a nanoHUB Sim2L, which extracts derived quantities and indexes all inputs and outputs in a FAIR data repository called ResultsDB. A separate nanoHUB workflow enables sequential optimization using active learning, where researchers define the optimization objective, and machine learning models are trained on-the-fly with all existing data, guiding the selection of future experiments. Inspired by the concept of ``frugal twin", the optimization task seeks to find the optimal recipe to combine food dyes to achieve the desired target color. With easily accessible and inexpensive materials, researchers and students can set up their own experiments, share data with collaborators, and explore the combination of FAIR data, predictive ML models, and sequential optimization. The tools introduced are generally applicable and can easily be extended to other optimization problems.

\end{abstract}

\section*{Introduction}
\label{introduction_section}

Digital twins (DTs) are virtual models of real-world systems that are continuously updated with experimental data to mirror the physical counterpart, enabling simulation-driven decision making and optimization. Modern DTs combine experimental data from the real-world system of interest with other sources of information, such as related experiments, physics-based simulations, and domain expertise. DTs also integrate machine learning tools to maximize their predictive power \cite{kalidindi2022digital}. In this context, data infrastructure that follows FAIR (findable, accessible, interoperable, and reusable) principles for data \cite{wilkinson2016fair} and workflows \cite{hunt2022sim2ls}is a central technology. Optimization and discovery tasks can benefit from access to relevant data, and geographically distributed collaborators can share data, analysis tools, and predictive models in real time. Platforms like nanoHUB \cite{strachan2010cyber} offer an opportunity to bridge this gap by supporting FAIR data and FAIR workflows to analyze, log and process data, enable machine learning and optimization tasks, and provide access to underlying computational resources.\cite{hunt2022sim2ls}.

This work demonstrates the use of nanoHUB's infrastructure for the development of a collaborative DT where independent researchers/labs contribute data to a shared repository, share analysis tools to process raw data, and share predictive tools to perform independent optimization tasks using active learning. The underlying infrastructure and tools are generally applicable and are demonstrated with a simple experiment using low-cost and widely available materials and a procedure that can be performed without domain-specific training. We took inspiration from the concept of a ``frugal twin" \cite{lo2024review}, a low-cost analog to a DT, where the experimental setup is simplified to capture only key behaviors of a more complex system, dramatically lowering barriers to entry.

Frugal twins are particularly valuable in the context of education, as they offer a simple and low-cost platform that captures the interplay between the physical and virtual worlds, iterative model updates, and decision-making \cite{macleod2022flexible}. Baird and Sparks developed a minimal self-driving laboratory (CLSLab), initially demonstrated with light as the medium of choice, showcasing the feasibility of accessible, autonomous experimentation. \cite{baird2022minimal} Subsequent studies emphasized the utility of such minimal systems as platforms for accessible education, prototyping, and the widespread adoption of self-driving laboratory methodologies.\cite{lo2024review,roch2020chemos} A team at Argonne National implemented an autonomous dye-mixing and color-matching system utilizing CMYB base dyes, prioritizing precision and internal system control, but sacrificing experimental accessibility. \cite{ginsburg2023exploring, vescovi2023towards}

Prior work in the field of frugal twins has emphasized low-cost hardware \cite{macleod2022flexible} and DT in the field of materials efforts have emphasized developments in predictive modeling \cite{slautin2024bayesian} and multi-step processes \cite{tao2018digital, szymanski2023autonomous}. This paper focuses on the role of shared, cloud-enabled data and computing infrastructure, an aspect critical for collaboration, reproducibility, and model improvement, which remains underexplored. We present the design and implementation of an open collaborative DT, demonstrate how shared data improves the predictive model and how it can accelerate optimization tasks even when collaborators have independent objectives. We emphasize the educational potential of this system and its value in research as a testbed to accelerate scientific discovery.

\section*{Methods}

We present a low-cost (under 10 USD) and accessible physical experiment in which users mix four dye colors in water to achieve a target color. Serving as a proof-of-concept, the setup enables broad participation in scientific experimentation with minimal equipment or technical expertise. The DT consists of experimental data and associated predictive models. The data includes pairs of dye recipes (defined by drop counts), their corresponding RGB color values, and a machine learning model trained on these data. As demonstrated below, the DT can be used to predict optimal recipes for user-defined target colors. All tools and data run and reside in the cloud and interfaces are accessible through the web so users do not have to download or install any software. 

\begin{figure}[h!]
  \centering
  \includegraphics[scale=0.75]
  {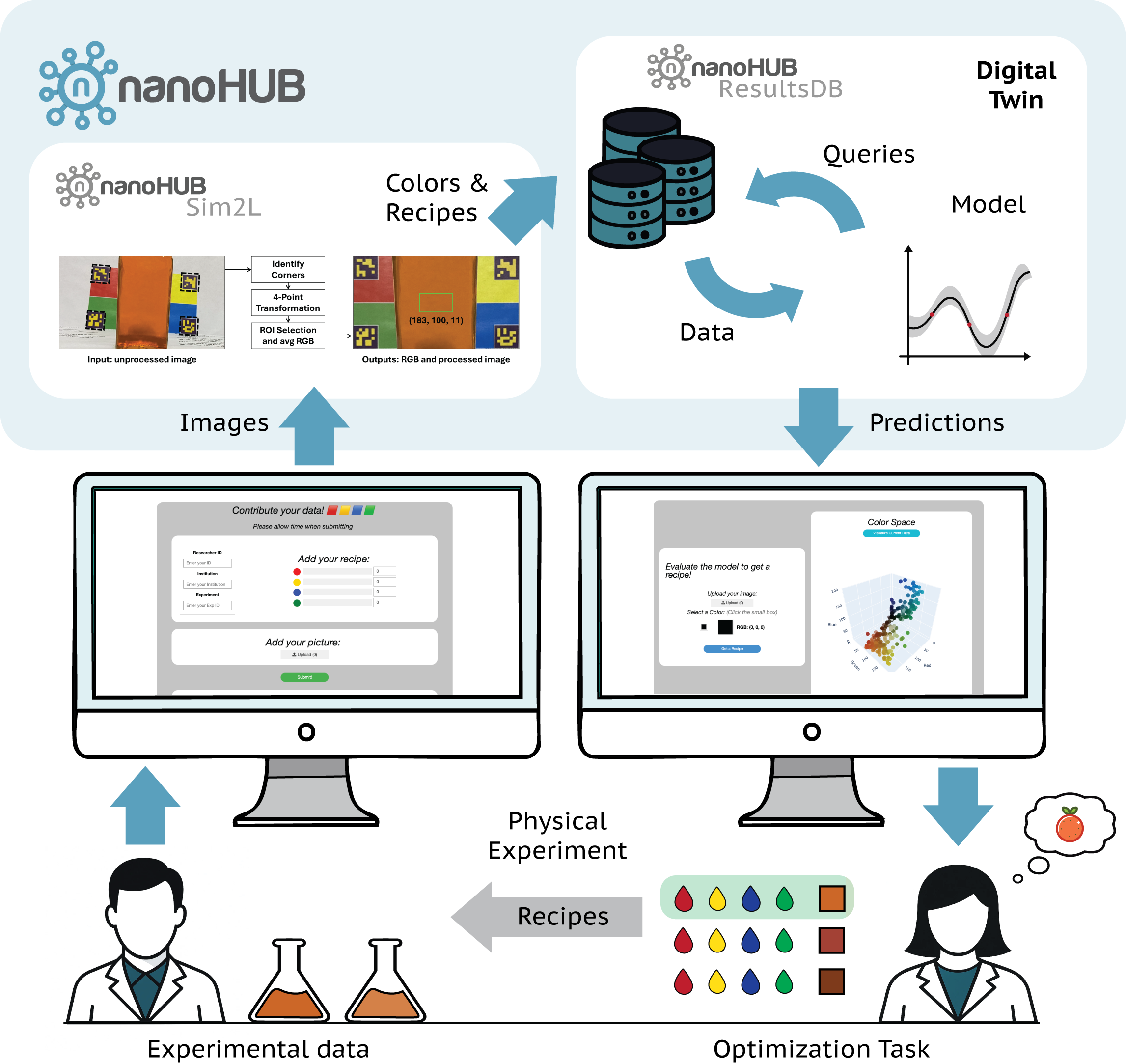}
  \caption{Experimental workflow enabled by the Active Learning GUI and sim2l. Users input a target color via RGB values or image upload. A trained model accesses nanoHUB’s ResultsDB and utilizing a Gaussian Process Regressor suggests both an optimized and exploration recipe based on expected improvement, which users can test and be uploaded to nanoHUB’s ResultsDB. Repeated recipes from the database are noted. This creates a continuous loop of experimentation, data contribution, and model refinement, supported by nanoHUB’s cloud infrastructure.}
  \label{fig:Overview}
\end{figure}

All stages of the workflow, from data ingestion and validation, to image processing, indexing, and model training, are implemented using nanoHUB's cloud infrastructure, as schematically illustrated in Figure \ref{fig:Overview}. The platform enables real-time model updates, centralized data logging, and asynchronous collaboration across users, made possible by its open-access tools and publicly available datasets. Our focus is on ensuring standardization, reproducibility, and iterative model development. We showcase a scalable and FAIR scientific infrastructure that is applicableacross disciplines and beyond this specific use case. The following paragraphs describe the main components of the DT and its interface with researchers.

\textbf{Interface.} Researchers and students interact with the DTs via two graphical user interfaces (GUIs): one to upload experimental data (\textit{Contribute Data}), and a second one to submit a target color and obtain optimal recipes (\textit{Evaluate Model}), both available online in the ``HueLogic" \cite{nH_HueLogic} tool, available in nanoHUB.

\textbf{Data generation and upload.} Users follow a simple procedure to mix food dyes in water using a basic dilution protocol, as referenced in the Supplementary Material, and capture the resulting color using a digital camera. Participants follow a standardized setup using a printable template under consistent lighting to ensure repeatable image capture. With their results at hand, they upload their picture, recipe, and metadata (contributor, institution) using the \textit{Contribute Data} GUI.

\begin{figure}[h!]
  \centering
  \includegraphics[scale=0.5]
  {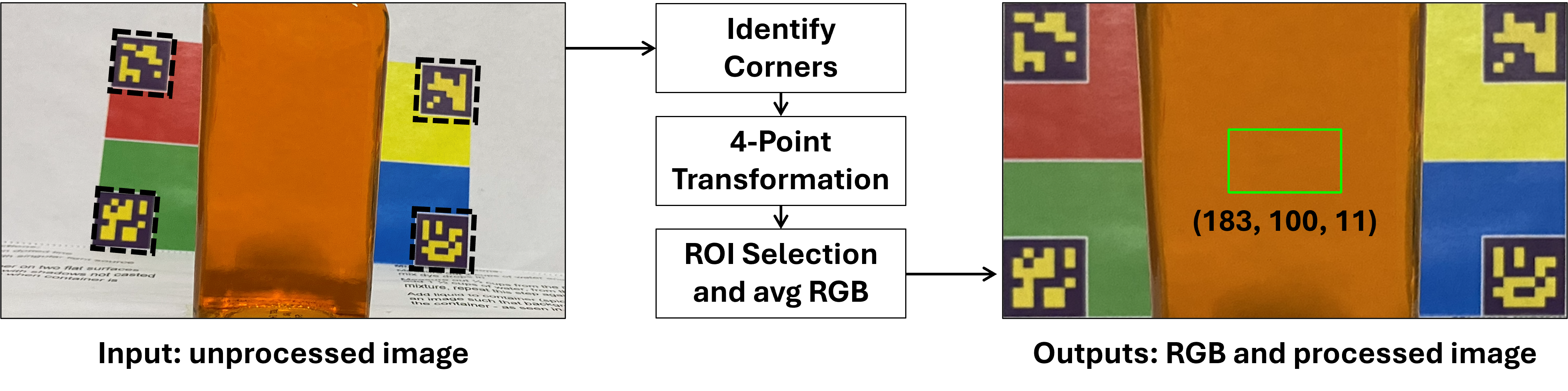}
  \caption{The nanoHUB-hosted Sim2L standardizes image processing by detecting ArUco markers, correcting geometry, extracting a central region of interest (ROI), and calculating its average RGB color. Results are stored in nanoHUB’s ResultsDB for reproducibility and machine learning use.}
  \label{fig:sim2l}
\end{figure}

\textbf{Data processing, indexing, and storage.} As with most experiments in science and engineering, raw data must be processed to extract meaningful results. We use nanoHUB’s Sim2Ls for data processing and ingestion. Sim2Ls are end-to-end workflows with declared inputs and outputs and have been used successfully in the past to implement computational workflows \cite{farache2022active, chen2025discovery}, as well as to ingest computational \cite{nykiel2023high} and experimental \cite{mishra2024mass} data. Sim2Ls have unique identifiers, are accessible online, and their services and requirements can be assessed via an API. The ``HueLogic" Sim2L \cite{nH_HueLogic} takes as input a picture of the printable template and the container with the solution, along with the recipe used and additional metadata. It identifies four reference ArUco markers \cite{garrido2014automatic} embedded in the printed template to locate the dye container and apply a geometric transform. A central region of interest (ROI) within the container is automatically highlighted and extracted. The average RGB color of the ROI is calculated using the OpenCV package \cite{opencv_library} and returned as a validated output. This is shown in Figure \ref{fig:sim2l}. This reproducible image processing workflow ensures consistency, minimizes user bias, and enables downstream use of the data for machine learning. Importantly, whenever a Sim2L is executed, all inputs and outputs are stored in nanoHUB’s ResultsDB. The dataset associated with this Sim2L is available in nanoHUB and can be queried using a standard API.

\textbf{Machine learning model.} The DT consists of the data and a predictive machine learning model. In this example, we use Gaussian Process Regressors (GPRs) to learn the mapping from input features (the number of drops of red, yellow, blue, and green dyes) to each of the output channels: R, G, and B. A separate GPR is trained for each output channel. Here, we define the distance metric as a squared norm to have a closed-form solution for the combined mean and uncertainty with the assumption that the models for each output channel are independent. The model is accessed via the \textit{Evaluate Model} GUI, where researchers can input a target color by uploading an image or entering RGB values. Upon submission, the model retrieves all relevant data available from nanoHUB's resultsDB, which can be optionally filtered, and trains a GPR on-demand using the selected dataset. Then, it evaluates all feasible recipes within the designed space (0-20 drops per dye) and returns two recipe suggestions: 

\begin{itemize}
  \item \textbf{Optimal recipe:} A recipe expected to match the target most closely, minimizing cumulative error across RGB channels.
  
  \begin{equation}
    x^* = \underset{x \in \mathcal{X}}{\arg\min} \; \bigl\|\mathbb{E}[M(x)] - y_{\text{target}}\bigr\|^2
  \end{equation}
  Where:
  \begin{itemize}
    \item \(x^*\) is the optimal recipe (a vector of drop counts for red, yellow, blue, and green).
    \item \(\mathcal{X}\) is the set of all feasible recipes (e.g., 0–20 drops per dye, indicating saturation.).
    \item \(\mathbb{E}[M(x)]\) is the GPR’s mean prediction of the RGB color for recipe \(x\).
    \item \(y_{\text{target}}\) is the user-defined target RGB color.
  \end{itemize}
  
  \item \textbf{Exploration recipe:} A recipe selected to improve the model by exploring uncertain regions, based on the Expected Improvement (EI) acquisition function \cite{jones1998efficient, vazquez2010convergence}.
  
  \begin{equation}
    \begin{split}
      x^* = \underset{x_i \in \mathcal{X}}{\arg\min} \;\; \rho\!\bigl(\mathbb{E}[M(x_i)] - \mathbb{E}[M(x_{\text{best}})],\,\sigma[M(x_i)]\bigr)\\
      \text{where}\quad
      \rho(z, s) =
      \begin{cases}
        s\,\phi'\!\bigl(\tfrac{z}{s}\bigr)\;+\;z\,\phi\!\bigl(\tfrac{z}{s}\bigr), & s > 0,\\
        \max(z,0), & s = 0.
      \end{cases}
    \end{split}
    \label{eq:MEI}
  \end{equation}
  Where:
  \begin{itemize}
    \item \(x^*\) is the selected exploration recipe.
    \item \(x_{\text{best}}\) is the recipe in the database that yields the lowest error.
    \item \(\mathbb{E}[M(x)]\) is the mean prediction of the GPR for the recipe \(x\).
    \item \(\sigma[M(x)]\) is the standard deviation predicted by the GPR (uncertainty).
    \item \(\phi(z)\) is the standard normal CDF and \(\phi'(z)\) its PDF.
  \end{itemize}
\end{itemize}

This setup enables a form of sequential, on-demand active learning. In our demonstration, the model is retrained every time a user queries it. For larger-scale systems, model training would typically occur offline and as new data are added. In our example, users can test the suggested recipe in the physical setup and, optionally, submit the result back to further train the model. The active learning loop continuously refines the model using real-world feedback and nanoHUB’s cloud infrastructure for validation, storage, and model access.


\subsection*{Results}

To evaluate the collaborative model, we compare the results of four researchers, each seeking to obtain a target color selected from sports teams and sharing data, with a single researcher optimizing a single color. In all cases, we began with a set of seven predefined “corner point” recipes, shown in Table \ref{initial_recipes}, to seed the initial dataset and support early interpolation by the active learning model. In the collaborative experiment, a single person simulated the four researchers and performed the experiments suggested by the active learning app (using the optimal recipe suggestion) in sequence and uploaded the results using the ``HueLogic" Sim2L \cite{nH_HueLogic}. The target colors for each simulated researcher are as follows:

\begin{itemize}
    \item Scientist 1 – Fever Yellow: (255, 213, 32)
    \item Scientist 2 – Giants Orange: (253, 90, 30)
    \item Scientist 3 – Cavaliers Red: (134, 0, 56)
    \item Scientist 4 – Dolphins Blue: (0, 142, 151)
\end{itemize}

\begin{figure}[h!]
  \centering
  \includegraphics[scale=0.4]
  {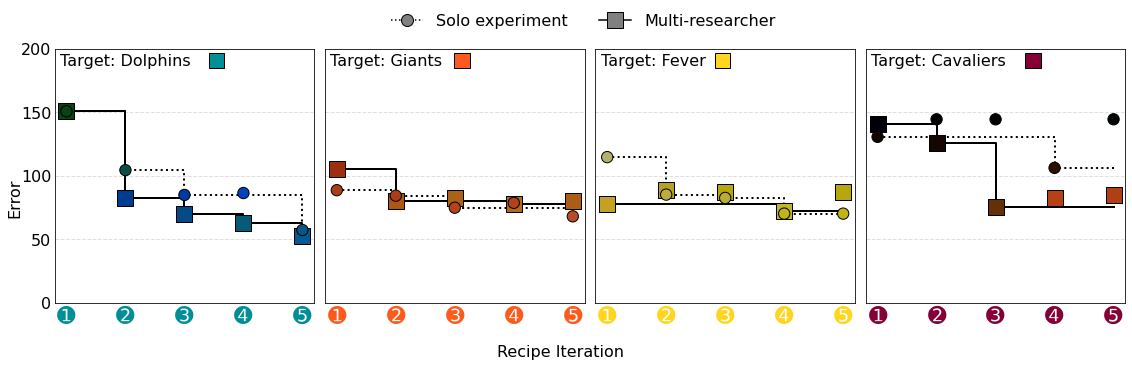}
  \caption{Active learning performance of solo- and multi-user experiments for different target colors. Lines represent the minimum cumulative error achieved up to the iteration. Dashed lines and circle markers represent the multi-researcher approach; solid lines and square markers represent the solo-researcher approach. Markers are colored to represent the color obtained in each experiment.}
  \label{fig:errors}
\end{figure}

With each successive recipe tested, the dataset grows by one data point, enabling subsequent predictions to be made using increasingly informative training data. By the time the active learning loop returned to the first researcher, the dataset had expanded to eleven samples, the seven original plus new ones. Figure \ref{fig:errors} compares the collaborative approach (squares) with the single-investigator approach (circles). We plot the error of the predicted color as a function of the optimization cycle. The error is defined as the Euclidean distance between the RGB vector of the experiment and that of the target. The target color is shown in the legend and the color of the symbol represents the actual color predicted by the ML model.

In all cases, the SDL was able to quickly find recipes that resemble all target colors. As will be discussed below, the target colors RGB values lie outside those achievable with our experimental setup, leading to a limit of minimal error. A sample-to-sample variability in color resulting in an error of 30 also contributes to the minimum error achievable. In all cases, sharing data, even when the objectives are different, either matches or outperforms individual efforts.

\begin{figure}[h!]
  \centering
  \includegraphics[scale=0.75]
  {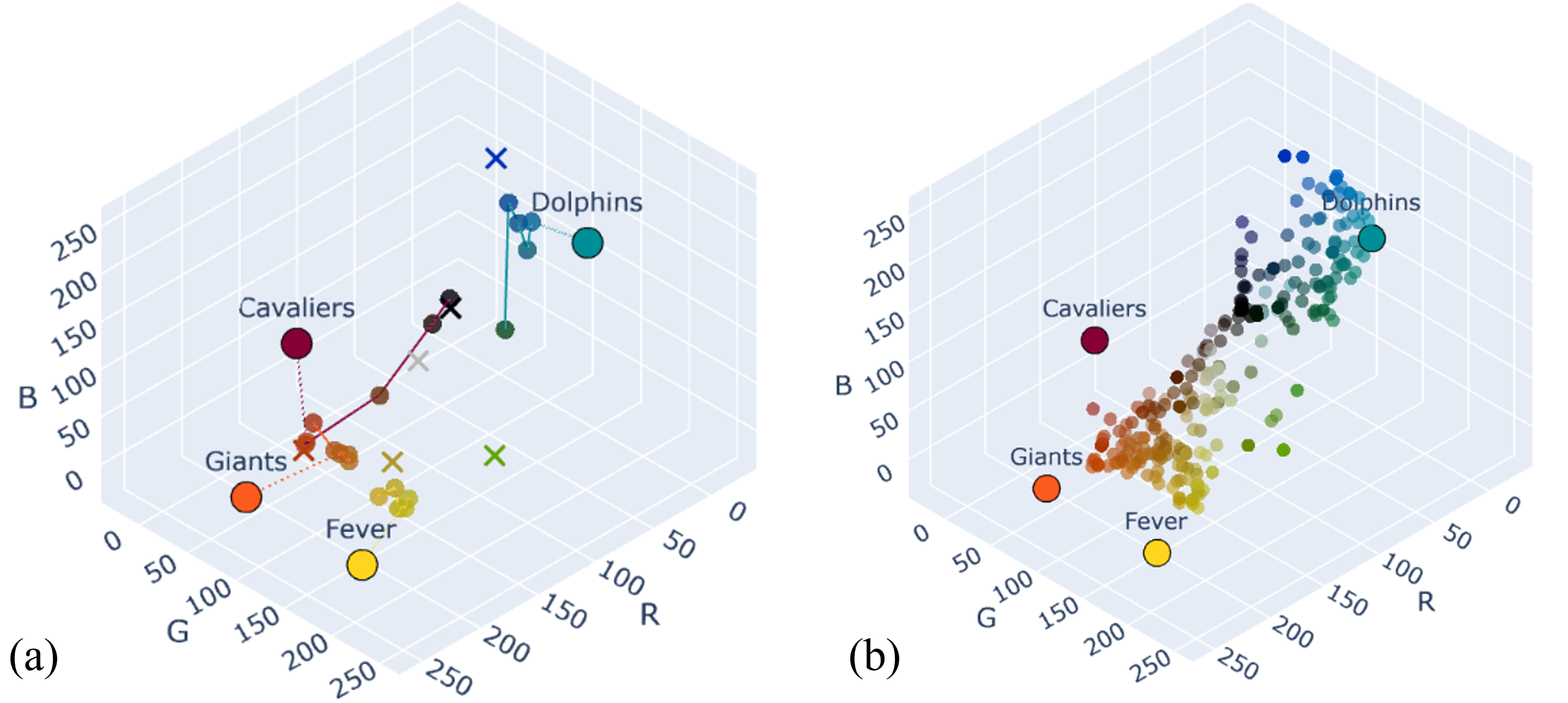}
  \caption{(a) Multi-user experiment testing paths for each target color. We observe that optimization paths for Giants orange and Cavaliers red intersect. Fever yellow and Dolphins blue remain in distinct regions. All optimizations leverage all training data. The "X" marks indicate initial training set, with dotted lines connecting Trial 5 points and the target color. (b) Color space representation showing all tested points, including those outside the specific experiment, highlighting the limitations of the dye capabilities. Target colors for Giants orange, Fever yellow, and Cavaliers red fall well outside the achievable color space within the experimental setup.}
  \label{fig:colorspace}
\end{figure}

To explore the optimization process performed by the sequential optimization algorithm, Figure \ref{fig:colorspace} (a) shows the colors selected by each optimization procedure in the collaborative approach. We observe crossing of paths between Giants orange and Cavaliers red, with Fever yellow and Dolphins blue optimizing within their own space, yet using all data for training. The initial training data points are displayed as "X" marks. Figure \ref{fig:colorspace} (b) highlights the limitations of the dye experiment. The markers in the figure represent a systematic exploration of the designed space by selecting the ``Optimal Recipes". We find that the target colors for Giants orange, Fever yellow, and Cavaliers red fall well outside the achievable color space within the experimental setup, with no recipes producing a RGB value greater than 200.

\section*{Discussion}
\label{discussion_section}

This work demonstrates how nanoHUB's compute and data infrastructure (Sim2Ls, ResultsDB, and cloud GUIs) can support collaborative, distributed SDLs and DTs with minimal need for local resources. The dye mixing experiment serves as a concrete example of how the FAIR infrastructure supports real-time model updates, shared validation pipelines, and controlled data sharing between users and locations. Our work highlights that accessible repositories like the ResultsDB, when combined with cloud computing and web-enabled GUIs, can enable asynchronous collaboration. Extensions of this work to distributed, interoperable, storage and compute will likely be needed for larger data sets. In this first demonstration, we used a per channel GPR model (one for R, G, and B) and its outputs are not constrained to lie in the 0 to 255 range. Future work should explore these modeling choices, including joint multivariate GPRs. 

We note that the experimental color space achievable with our procedure was notably constrained, with several target RGB values (particularly for Fever yellow, Giants orange, and Cavaliers red) falling outside the range of achievable outputs. A primary limitation is the spectral range and absorption properties of the food dyes used, as many vivid or saturated targets remain inaccessible. Additional base dyes, such as CMYB \cite{ginsburg2023exploring, vescovi2023towards}, would expand the achievable color space.

We note that while the experiments for Dolphins Blue, Giants Orange, and Fever Yellow show little or no improvement between the solo and multiuser experiments, Cavaliers Red showed significant improvement with the collaborative approach. This is likely a result of the overlap between the Giants Orange and Cavaliers Red explored space, allowing for increased training for Cavaliers Red, where the other target colors and their recipes are isolated. The absence of external calibration in our analysis script (e.g., color swatches or grayscale cards) limits the ability to correct for lighting and exposure conditions across setups, yet the template provides standard colors to allow for future color correction of images if desired. Image capture using participant devices introduces variation in white balance, sensor sensitivity, resolution, and postprocessing, all of which can distort the recorded color from the actual mixture. This is further discussed in the supplementary material in the Repeatability section. The Sim2L currently applies transformations and extracts a single average color per image. Future workflows could incorporate color gradients or confidence intervals to better characterize experimental uncertainty.

\section*{Conclusions}
\label{conclusions_section}

We demonstrated a collaborative DT using nanoHUB's open infrastructure, where geographically distributed researchers can share data, analysis scripts, and predictive machine learning models. We implemented a low-cost experiment using common materials and a simple procedure that can be easily replicated for training and education purposes. Through active learning, we showed that the DT improves progressively with each new data point, guiding users toward optimal recipes. Despite the limitations of our experimental setup, dictated by our aim of simplicity and low cost, the example can be used to introduce students and researchers to the concepts of DTs, FAIR data, predictive models, and sequential learning. The approach and infrastructure are generally applicable, and the GUI apps, the ingestion tool, and the machine learning models can be easily adapted to serve other applications. We believe that this general framework, grounded in FAIR principles and cloud-based infrastructure, offers a scalable path toward accelerating discovery and optimization across a wide range of scientific domains.

\section*{Acknowledgements}

This study was funded by the US National Science Foundation through the FAIROS program (Award 2226418). The funder played no role in study design, data collection, analysis and interpretation of data, or the writing of this manuscript. We acknowledge computational resources from nanoHUB and Purdue University through the Network for Computational Nanotechnology.

\section*{Code and Data availability}
\label{data_availability_section}

The underlying code for both graphical user interfaces, the Sim2L\cite{hunt2022sim2ls} workflow for image processing, and the dataset generated for this study are all part of a single project implemented on nanoHUB\cite{strachan2010cyber} and available online\cite{nH_HueLogic}.
 
\section*{Ethics declarations}

\subsection*{Competing interests}

All authors declare no financial or non-financial competing interests. 

\subsection*{Author Information}

\textbf{Corresponding Author}

\textbf{Alejandro Strachan} - School of Materials Engineering and Birck Nanotechnology Center, Purdue University, West Lafayette, Indiana 47907; Email: strachan@purdue.edu \\

\bibliographystyle{unsrt}
\bibliography{references.bib}

\newpage

\section*{Supplementary Material}

\subsection*{Training Set}

\begin{table}[ht]
\caption{Initial training recipes used for early training by the active learning model. Each recipe specifies drop counts for red, yellow, blue, and green dyes.}
\label{initial_recipes}
\vskip 0.3in
\begin{center}
\begin{tabular}{cccc}
\toprule
\textbf{Red Drops} & \textbf{Yellow Drops} & \textbf{Blue Drops} & \textbf{Green Drops} \\
\midrule
 0  & 0  & 0  & 0  \\
 20 & 0  & 0  & 0  \\
 0  & 20 & 0  & 0  \\
 0  & 0  & 20 & 0  \\
 0  & 0  & 0  & 20 \\
 10 & 10 & 10 & 10 \\
 20 & 20 & 20 & 20 \\
\bottomrule
\end{tabular}
\end{center}
\vskip -0.1in
\end{table}

\subsection*{Dilution}

The mixing and dilution process maximizes the achievable color space while minimizing resource consumption. By initially mixing dye drop quantities into 2 cups of water, a consistent and reproducible starting concentration is established. The subsequent dilution steps, removing half a cup of the liquid and adding one and a half cups of water to the removed mixture, repeated twice, allow for a controlled reduction in dye concentration. This approach conserves water, using a total of 5 instead of 32 cups of water, while enabling the exploration of a broader range of lighter hues that are difficult to reach and maximizing the number of dye drops used to optimize resolution between recipes.


\begin{figure}[h!]
  \centering
  \includegraphics[scale=0.5]
  {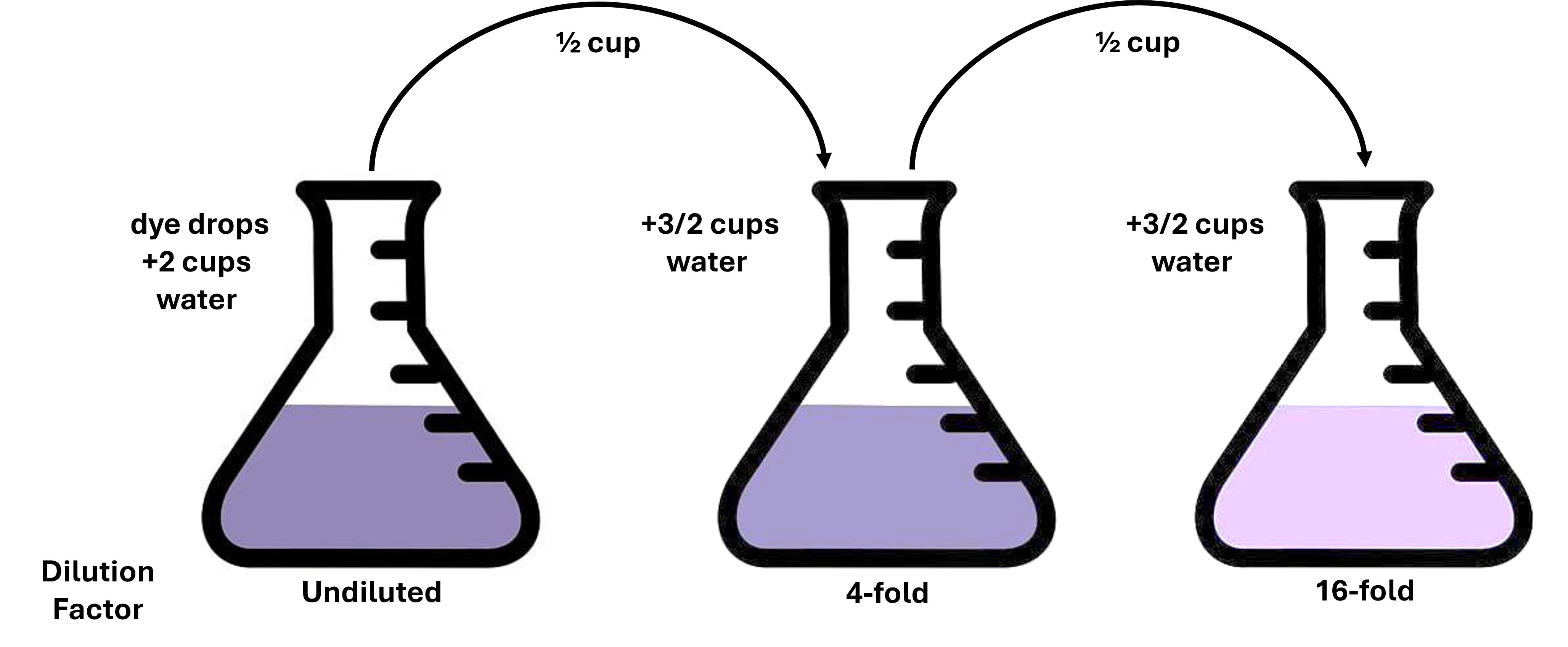}
  \caption{The mixing and dilution process provides the largest color space while conserving drop and water usage. Mix the recipe dye drop quantities with 2 cups of water. Then, remove 1/2 cup and add 3/2 cups of water to the removed mixture. Repeat this process once more.}
  \label{fig:Dilution}
\end{figure}

\subsection*{Repeatability}

\begin{table}[ht]
\caption{Observed RGB variance from repeated image captures of the same mixtures under identical lighting conditions. Each entry corresponds to the Trial 5 recipe from the demonstration experiment, highlighting the variability introduced by image capture noise and processing steps.}
\label{rgb_trials}
\vskip 0.3in
\begin{center}
\begin{tabular}{lcccc}
\toprule
& \textbf{Dolphins (RGB)} & \textbf{Giants (RGB)} & \textbf{Fever (RGB)} & \textbf{Cavaliers (RGB)} \\
\midrule
Picture 1 & (4, 89, 149) & (182, 95, 23)  & (188, 165, 34) & (192, 70, 21) \\
Picture 2 & (4, 92, 156) & (188, 98, 24)  & (193, 171, 37) & (177, 65, 20) \\
Picture 3 & (4, 90, 151) & (190, 99, 24)  & (192, 169, 35) & (179, 65, 20) \\
Picture 4 & (4, 90, 153) & (182, 95, 23)  & (191, 167, 35) & (176, 64, 19) \\
Picture 5 & (4, 90, 152) & (188, 98, 24)  & (188, 166, 35) & (177, 65, 19) \\
\midrule
Sample Variance & (0, 1, 7) & (14, 4, 0) & (5, 6, 1) & (45, 6, 1) \\
\bottomrule
\end{tabular}
\end{center}
\vskip -0.1in
\end{table}

\end{document}